\documentclass[runningheads]{llncs}
\usepackage{graphicx}
\usepackage{xcolor,colortbl}
\usepackage{amsmath,amssymb} 
\usepackage[margin=0pt]{subfig}
\usepackage{overpic}
\usepackage{color}

\def\vec#1{\mathchoice{\mbox{\boldmath $\displaystyle\bf#1$}}
{\mbox{\boldmath $\textstyle\bf#1$}}
{\mbox{\boldmath $\scriptstyle\bf#1$}}
{\mbox{\boldmath $\scriptscriptstyle\bf#1$}}}
\def\vaz#1{\protect\vec #1}

\def\mat#1{\mathchoice{\mbox{\boldmath$\displaystyle\tt#1$}}
{\mbox{\boldmath$\textstyle\tt#1$}}
{\mbox{\boldmath$\scriptstyle\tt#1$}}
{\mbox{\boldmath$\scriptscriptstyle\tt#1$}}}
\def\m#1{\protect\mat #1}

\newcommand{\lra}{\leftrightarrow}

\newcommand{\figref}[1]{\Fig~\ref{#1}}
\newcommand{\secref}[1]{Section~\ref{#1}}

\newcommand{\tabref}[1]{Table~\ref{#1}}

\definecolor{Gray}{gray}{0.85}
\definecolor{LightCyan}{rgb}{0.88,1,1}
\definecolor{LightRed}{rgb}{1,0.88,1}
\definecolor{LightGreen}{rgb}{1,1,0.88}

\newcolumntype{P}[1]{>{\arraybackslash}p{#1}}
\newcolumntype{a}{>{\columncolor{LightCyan}}c}
\newcolumntype{b}{>{\columncolor{LightRed}}c}
\newcolumntype{d}{>{\columncolor{LightGreen}}c}

\makeatletter
\newcommand{\corr}{{\vaz x_i^t}}
\newcommand{\src}{{{\vaz x_i^s}}}
\newcommand{\srcthreed}{{\vaz X_i}}
\def\onedot{.}

\def\srcview{{\em source} view}
\def\tgtview{{\em target} view}
\def\sculpdata{{Sculpture} dataset}

\def\scanned{Scanned}

\def\Fig{Fig\onedot}

\begin{document}
\pagestyle{headings}
\mainmatter
\title{3D Surface Reconstruction by Pointillism} 

\titlerunning{Pointillism}

\authorrunning{O. Wiles and A. Zisserman}

\author{Olivia Wiles and Andrew Zisserman}

\institute{Visual Geometry Group,\\
	Department of Engineering Science, University of Oxford\\
	\email{ \{ow,az\}@robots.ox.ac.uk}
}

\maketitle

\begin{abstract}
The objective of this work is to infer the 3D shape of an object from a
single image. We use sculptures as our training and test bed, as these have
great variety in shape and appearance. 

To achieve this we build on the success of multiple view geometry (MVG)
which is able to accurately provide {\em correspondences} between
images of 3D objects under varying viewpoint and illumination
conditions, and make the following contributions: first, we introduce
a new loss function that can harness  image-to-image
correspondences to provide a supervisory signal to train a deep
network to infer a depth map.  The  network is trained end-to-end
by differentiating through
the camera. Second,  we develop a processing pipeline to automatically  generate
a large scale  multi-view set of correspondences  for training the network. 
Finally, we demonstrate that we can indeed obtain a depth map of a novel object from a single image
for a variety of sculptures with varying shape/texture, and
that the network generalises at test time to new domains (e.g.~synthetic images).
\end{abstract}

\begin{figure}[t]
\centering
\includegraphics[width=\linewidth]{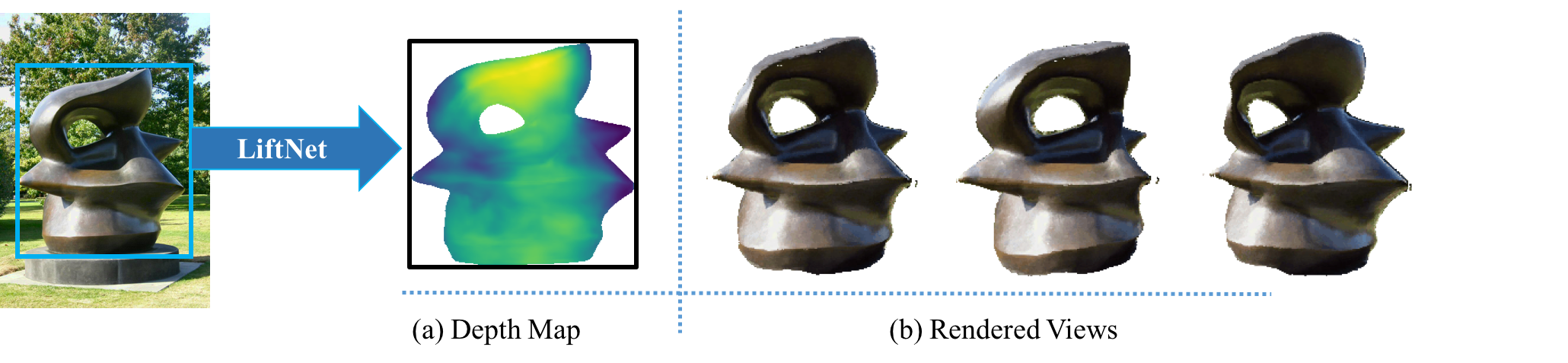}
\caption{
Given this image of the \emph{Spindle Piece} by {\em Henry Moore},  LiftNet  predicts the 3D shape of the sculpture via a depth map (a), where blue is further away and yellow nearer,  and thereby enables new views to be rendered (b).
 LiftNet is trained self-supervised on real images using correspondences \emph{without} any knowledge of depth or camera parameters. 
}
\label{figure:teaser}
\end{figure}

\section{Introduction}
\label{sec:introduction}

Humans are able to effortlessly perceive 3D shape of a previously unseen object from a single image
-- or at least we have the impression that we do this.
For example for a piecewise smooth sculpture such as the one by
Henry Moore in \figref{figure:teaser}, we know where there are 
concavities, convexities and saddles,  as well as where there are holes and sharp points.
{\em How} this is achieved has long been studied in computer vision in terms of geometric cues
from the silhouette~\cite{Koenderink84}, the texture~\cite{Witkin81_mine,Malik97_mine,Blake90_mine}, self-shadows, specularities~\cite{Fleming04}, shading~\cite{Zhang99_mine,Barron15}, chiaroscuro~\cite{Koenderink80}, etc.

In this paper our objective is to be able to reconstruct such objects
from a single image. Deep learning has significantly boosted progress in 3D
reconstruction from single images, but so far methods have mostly depended on the availability of
synthetic 3D training examples, or using a single class, or pre-processing the data
using SfM and MVS to extract depth.  In contrast, 
our self-supervised approach is to learn
directly from real images, capitalizing on many years of research 
on MVG~\cite{Hartley04a,Lowe04,Snavely06,Schaffalitzky02}  that is able to automatically determine
matching views of a 3D object and generate point correspondences,
without requiring any explicit 3D
information as supervision.

The key idea is to use image-to-image point correspondences to provide
a {\em training} loss on the depth map predicted by a 
CNN, called {\em LiftNet}.
This is illustrated in
figure~\ref{figure:correspondences}. 
Suppose we are attempting to infer the depth of the object
in a \srcview, and there are a number of image point
correspondences available between the \srcview \ and a \tgtview \ (where a correspondence is defined by the projection of a
3D surface point  into the source and target  views). 
A correspondence can be computed in two ways.
First, it can be computed using matching methods from 
MVG (such as SIFT, and epipolar geometry). This method does not involve using
the depth of the surface and we treat these correspondences as ground truth.
Second, it can be computed by inferring the depth of the point in the \srcview \ and projecting
the 3D point into the \tgtview. 
If the CNN correctly predicts the depth of the points in the \srcview, then
the projected points will coincide with the ground truth correspondences in the \tgtview; however, 
if the the depth prediction
is incorrect, then the distance between the projected and corresponding points -- the re-projection error -- defines a  loss that can be used to train the network.

Of course, the correspondences between two views of a particular
sculpture only provide constraints at those points on the surface --
and correspondences will mainly be found at surface texture, surface
discontinuities, and boundaries~\cite{Arandjelovic12a_mine}, i.e.\ not uniformly across
the surface. However, for each sculpture there are multiple pairs of
images; and each pair can `probe' (and constrain) different points on the
surface according to its correspondences.  
Finally, and most importantly, the network must learn to predict correspondences  
not just for a particular sculpture, but
for {\em all} the sculptures (and all their view pairs) in the training set -- and we have
170K training pairs and around  31M training correspondences. The only way it can solve this task
is to infer 3D shape for each image.

To this end, we formulate a new deep learning framework for extracting
3D shape which is similar to the artistic \emph{pointillist} style.
Analogously to how pointillists build up colour variation in an image
from dots of discrete colour, we use points in correspondence between
images of an object in order to train a network over time to learn the
3D shape of the object.  

\noindent \textbf{Contributions.} 
This work presents three contributions: first,
  to use corresponding points to formulate a
differentiable loss on the object shape that can be used to train a
network from scratch (\secref{sec:approach}). The formulation includes differentiating through
the camera to train the network end-to-end.

The second contribution is a pipeline based on MVG for automatically extracting robust correspondences between
multiple pairs of images of a sculpture (\secref{dataset:collection}).  
We use
these correspondences to train the network on real images,
{\em without} ground truth 3D depth information.  This is done
entirely automatically and is the first system to our knowledge to
learn to predict shape end-to-end for a set of objects by using
correspondences and geometry in this manner.  

The final contribution is our experimental results in~\secref{sec:experiments}, which demonstrate that the trained network can not only predict depth for the given domain but also generalises to synthetic data, allowing its generalisation capability to be evaluated quantitatively.

\section{Related Work}

\paragraph{Depth Prediction.}
The ability to learn depth using a deep learning framework was introduced by~\cite{Eigen14}, who use a dataset of ground truth depth and RGB image pairs to train a network to predict depth.
This has been improved on with better architectures in~\cite{laina16,Eigen15} and generalised to ordinal  relationships in~\cite{Chen16,Zoran15}.

A recent set of works have considered how to extract the 3D depth of a
scene between pairs of images without knowing the camera motion or
depth~\cite{Vijayanarasimhan17,Godard17,Zhou17,Ummenhofer16}.  This is
done by predicting both depth and cameras in the network.  This
information is then used to transform one view and the photometric error between the generated image and the ground truth is used to
train the network.  These works require that the two images be very
similar, such that the photometric error 
gives a robust and
sensible loss.  As a result, the images come from stereo datasets or
consecutive  video frames, such that the relative appearance  change is small.
On the other hand, our approach uses point correspondences
directly, and consequently  the images can vary dramatically in illumination,
texture, size, position, etc.~and our loss is robust to these changes.

\paragraph{3D Shape Prediction.} 
Going beyond depth prediction, which is view based, the entire 3D
shape of the object can be reconstructed from multiple views 
by using strong supervision from the known 3D geometry to predict
a voxel~\cite{Choy16,Girdhar16} or
point cloud~\cite{Fan16,Su17,Wu16} representation. Alternatively, 
the supervision can be from photo consistency or
silhouette constraints
\cite{Soltani17,Tulsiani17,Rezende16,Yan16,Gadelha16}.  However, these
methods require knowledge of the camera parameters in order to enforce
the geometric constraints.

These methods have been extended to deal with natural images in the
work of \cite{Zhu17,Novotny17,li2018megadepth}, but \cite{Zhu17} still
requires a synthetic dataset on which to train their network which is
then fine-tuned on real images.  \cite{Novotny17} uses
structure from motion (SfM)/multi-view stereo (MVS)~\cite{Hartley04a}
from a video sequence as the ground truth 3D shape on which to train
their network for reconstructing a finite set of classes; 
\cite{li2018megadepth} extends this idea to unordered image collection
of historic landmarks by using many images of the given landmark.  In
our case, we are not restricted to a finite set of classes,
and do not require a video sequence or many images of the same scene in order to
obtain a dense reconstruction, but instead train from the available correspondences directly, 
and these correspondences only need to exist over a handful of images.
As a result, our approach can be used with far fewer samples of each landmark or sculpture.

\begin{figure}[t]
\centering

\includegraphics[page=9,width=.83\linewidth]{./pclloss2}

\caption{An illustration of the  training  loss: $\mathcal{L}_{\text{corr}}$. Given {$\src \lra \corr$} and the best camera $\m P$, we minimise the error between {\bf$\m P${$\srcthreed$}} and {\bf{$ \corr$}}. The depth $d_i$ value of {$\srcthreed$} is LiftNet's predicted depth for the \srcview \ at {\bf{$ \src$}}. 
If $d_i$ were correctly predicted by LiftNet there would be no error as $\srcthreed$ would project to {$ \corr$}; the image 
distance between the  projected point and {$ \corr$} provides the training loss. As the network's prediction improves, the distance reduces to zero.
}
\label{figure:correspondences}
\end{figure}

\section{Approach}
\label{sec:approach}

The goal  is to recover 3D structure from a single image by predicting a depth map, 
but \emph{without} requiring ground truth 3D information in training.
In this section we first define the loss functions used to train the network.
Then the  LiftNet architecture is described in \secref{sec:architecture}.
In the following we assume that correspondences between images are available (as
described in~\secref{dataset:collection}).

As introduced in \secref{sec:introduction}, the depth predicted by the
LiftNet CNN in the \srcview \ is supervised by using point
correspondences as follows: (i) let the set of correspondences be
denoted as $\src \lra \corr$, where $ \src $ are the points in the
\srcview, and $ \corr $ the points in the \tgtview.  (ii) Then
in the \srcview \ we can determine the 3D points $ \srcthreed $ that
project to $ \src $ (since the network gives the depth of each point).
(iii) Since we know the correspondence between $ \srcthreed $ and
$\corr$ we can compute the best camera that projects the 3D points $
\srcthreed $ into the \tgtview. (iv) If the 3D shape has been
predicted perfectly, then the 3D points $ \srcthreed $ will project
perfectly onto $\corr$. If they do not, then this {\em reprojection
error} provides a loss that can be minimized to train the network.
The resulting loss is defined as:
\begin{equation}
\label{eq:loss}
	\mathcal{L}_{\text{corr}} = \frac{1}{N} \sum_{i=1}^N  d_R ( \m P \vaz X_i, \vaz x_i^t ) ; 
\end{equation} 
where $d_R ( . , . )$ 
denotes the Euclidean ($L_2$) pixel distance between vectors subject to a robustness function $R$.

This loss is a useful constraint, as it enforces important properties of the object, such as concavities and convexities. 
Moreover, this can be done for any pair  of images for which correspondences can be obtained. 
There is no requirement that the images be photometrically consistent -- e.g.~lighting, texture, position etc.~can vary dramatically between views.

Finally, a robustness term $R$ is added (\secref{sec:robustness}), as the 2D correspondences may be noisy (as explained in \secref{dataset:collection}).

\subsection{Point Correspondence Loss $\mathcal{L}_{\text{corr}}$} 
\label{sec:correspondence}

We minimise the projection error between $\srcthreed$ and $\corr$ using the best camera $\m P$:
$\frac{1}{N} \sum_{i=1}^N d_R ( \m P \vaz X_i, \vaz x_i^t )$.
The steps are as follows:

\paragraph {\bf A. Choose the camera.} This work assumes an affine camera and an orthogonal coordinate system in the \srcview,  which is why $\srcthreed = [x_i^s, y_i^s, d_i, 1]^T$ projects to $\src = (x_i^s, y_i^s)$ in the \srcview.
As has been noted previously~\cite{Hartley04a,Hong16,Hong17}, the affine case is a
very stable and useful approximation to perspective projection.
The reader is referred to the supplementary material for a detailed review of this camera model. 
However, we note that the ideas presented here (e.g.\ the method of differentiating the camera) generalise in a straight forward manner to the perspective case.

\paragraph {\bf B. Determine the camera.} 
We first determine the camera matrix $\m P$ by solving the
system of equations $ \corr $ $ = \m P $ $ \srcthreed $ for $\m P$. 
We know which values $\srcthreed$ and $\corr$ should correspond because LiftNet's prediction is simply a depth map, so $\src = (x_i^s, y_i^s)$ in pixels maps to $\srcthreed = [x_i^s, y_i^s, d_i, 1]^T$ ($d_i$ is the depth prediction at that point) and we know, via the correspondences, that $\src \lra \corr$, so $\srcthreed$ maps to $\corr$.
This gives the following system of equations:
\begin{equation}
\begin{bmatrix}
x_1^t & \dots & x_N^t \\
y_1^t & \dots & y_N^t \\
\end{bmatrix}
=
\m P
\begin{bmatrix}
x_1^s & \dots  & x_N^s \\
y_1^s & \dots & y_N^s \\
d_1 & \dots & d_N \\
1 & \dots & 1 \\
\end{bmatrix}.
\label{eq:systemofequations}
\end{equation}
However, directly solving the system of equations would be problematic due to the effect of outliers, (e.g.~noise in the data). 
A standard approach to deal with noise is to make use of RANSAC~\cite{Fischler81}.
This method solves a system of equations by finding a solution that satisfies the most constraints.
The satisfied constraints are called inliers, the others outliers.
In our case, we want to find $\m P$ such that the maximum number of pairs $\srcthreed$ and $\corr$ satisfy the condition $|\corr - \m P \srcthreed |_2 < T$ for some threshold $T$.
 Given the set of inliers $\vaz X_{i_{\text{inliers}}}, \vaz x_{i_{\text{inliers}}}^t$, a new system of linear equations is constructed: $\vaz x_{i_{\text{inliers}}}^t = \m P \vaz X_{i_{\text{inliers}}}$.

\paragraph {\bf C. Compute the loss.} Given $\m P$, all points $\vaz X_i$  are projected into the \tgtview \ and the error between their projection and known location  $\corr$ is computed.
The loss is then as given in~\eqref{eq:loss}.

\paragraph {\bf D. Differentiate through the camera.} 
In order to train the network end to end, it is necessary to compute the derivative $\frac{\partial \m P}{\partial d_i}$.
To do this, we re-write the system of equations such that $\m P$ is explicitly a function of $x_{i_{\text{inliers}}}^t$/$\vaz X_{i_{\text{inliers}}}$ such that computing the
derivative is straightforward.
 For ease of notation, the matrix of inliers $\vaz X_{i_{\text{inliers}}}$ is referred to as $\m X$ and of inliers $\m x_{i_{\text{inliers}}}$ as $\m x$ 
 from now on.
The pseudo-inverse $\m  X^+$  is computed using the singular value decomposition (SVD)~\cite{Strang1980}.
(If the SVD of a matrix $\m A$ is  $\m A = \m U \m \Sigma \m V^T$ then its pseudo inverse can be written as $\m A^+ = \m V \m \Sigma^+ \m U^T$.)
Then the system of equations can be re-written as:
\begin{align}
\m x &= \m P \m X \\
\m x^T &= \m X^T \m P^T  \\
(\m X^T)^+ \m x^T &\approx (\m X^T)^+ \m X^T \m P^T \approx \m P^T  \\
(\m V\m \Sigma^+ \m U^T) \m x^T &= \m P^T
\end{align}
Note that because the system of equations in \eqref{eq:systemofequations} is over-constrained then of course this is not an exact solution, but the pseudo-inverse solves the system of equations in the least-squares sense~\cite{Strang1980} which is what we require.
Also, note that because the outliers are ignored when computing $\m P$ due to RANSAC, we can ignore them in this computation and only consider the inliers.
This gives the forward pass.

To perform the backward pass, it is necessary to compute the derivative $\frac{\partial \m P}{\partial d_i}$.
This is a straight-forward application of the product and chain rule, except for the computation of the SVD.
However, previous work (e.g.\ \cite{Papadopoulo00}) has demonstrated how to compute these derivatives.
As a result we can back-propagate through the computation of $\m P$ to the estimation of the height values $d_i$.
This is achieved in practice using standard layers in a neural network library (e.g.~PyTorch~\cite{paszke2017automatic}).
Note that computing the gradients for $\m U$, $\m V$ could lead to potential instability if $\m X^T$ is not full rank or has repeated singular values; however, this was not a problem in practice.

\paragraph {\bf Discussion.} We note that our method 
computes $\vaz d$ up to an overall affine ambiguity.  This amounts to
a scaling and shearing in the depth prediction.  This ambiguity is
seen in human vision, as humans have been shown to reconstruct objects
(such as vases) up to an affine transformation in
depth~\cite{todd2004visual,koenderink1992surface}.  It is hypothesised
that this difficulty arises from the fact that, assuming Lambertian
reflectance and given a single image, the surface of an object can
only be recovered up to a generalized bas-relief
ambiguity~\cite{belhumeur1999bas}.

\subsection{Robustness}
\label{sec:robustness}
As the correspondences and segmentations will be noisy, it is necessary that the 
loss function is robust to these errors.  To do this we use a
smooth function to weight the errors~\cite{Liwicki16} so that errors
above a threshold $\tau$ are given a constant cost:
$
	\mathcal{R}(x) = 
	\begin{cases}
		\frac{1}{2} x^2 (1 - \frac{x^2}{2 \tau^2}) , & \text{if }  x^2 \leq \tau^2 \\
		\tau^2 / 4, & \text{otherwise}
		
	\end{cases}
$.

\subsection{Architecture}
\label{sec:architecture}

The architecture used is based on the U-Net \cite{Ronneberger15} variant of pix2pix \cite{Isola16}.
This architecture includes skip connections in order to maintain high level detail.
However, we incorporate two modifications.
First the last activation is replaced by a tanh layer to enforce that the output  lies between $[-1,1]$.
We impose this range so that the predicted depth does not grow too large, making training unstable.
As LiftNet learns depth up to a scaling factor in depth, this in no way constrains the types of surfaces that LiftNet can describe. 
Second, the nearest neighbour upsampler is replaced by a bilinear upsampler.
This mitigates against pixelated effects~\cite{Odena16,Chen17}.  
Please refer to the supp.~material for 
full  details.

\section{The Sculpture Dataset}
\label{dataset:collection}

\begin{figure*}[t]
\centering

\subfloat[The images are organised by artist (solid lines) into clusters (dashed lines). Note  the variety and complexity of the sculptures: 
in shape,  materials, lighting, and viewpoint.]{
\includegraphics[width=\linewidth,page=1]{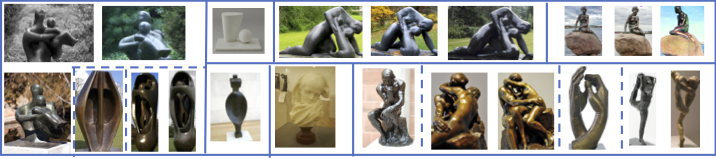}}

\subfloat[Sample correspondences for pairs of images.
The images may be taken at different times of year, in different contexts/illumination conditions and the material itself may change over time due to weather.]{
\includegraphics[height=0.16\linewidth]{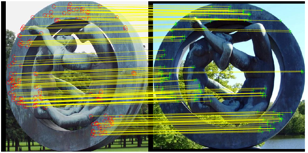} \qquad
\includegraphics[height=0.16\linewidth]{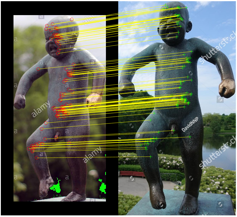} \qquad
\includegraphics[height=0.16\linewidth]{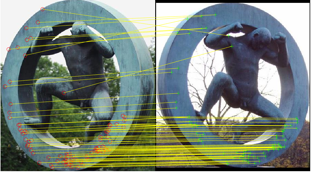}
\label{fig:samplecorrespondences}
}

\caption{{\bf The \sculpdata.} 
Note that this is only a tiny
subset of the clusters and a fraction of the
number of images within a cluster. Please refer to the supp.~material for more examples. 
}
\label{fig:samplesculptures}
\end{figure*}

We assemble a large scale dataset of images of sculptures for training and testing by combining
multiple public datasets~\cite{Arandjelovic11_mine,Arandjelovic12a_mine,Fouhey16,Knapitsch2017} and downloading additional images from the web.
The dataset  incorporates  a wide variety of artists, styles and materials.
It is divided at the artist level to prevent any information bleeding between the sets.
\tabref{table:sculpturesstats} gives the number of artists and works used as well as the train/val/test splits.

The dataset includes multiple works (sculptures) by different artists (sculptors) 
organised into a set of clusters.  
Within a cluster, the images are of the same sculpture (shape), but there may be multiple instances of the sculpture,
some made of different material. 
The utility of the dataset is that within a cluster
there are many point correspondences between image pairs that can be used for training 
the network. \figref{fig:samplesculptures} shows a sample of sculptures, correspondences
and an example cluster.  

\begin{table}[t]
\centering
\caption{Dataset statistics for the \sculpdata. Note the large number of artists and works. This results in a large variety of styles and shapes that LiftNet must contend with.
A total of 31M point correspondences ($\approx 181$ correspondences $\times$ $169K$ pairs)
are automatically generated and used to train LiftNet.}
\scriptsize
\begin{tabular}{| l  || c || c || c  || c || }
\hline
& Train & Val  & Test & All \\ \hline 
\#Artists & 138 & 7 & 1 &  143 \\
\#Works & 1031 & 27 & 129 & 1187 \\
\#Matching Pairs & 168726 & 552 & 13166 & 182K \\
Avg \# Correspondences per Pair & 181 & 223 & 174 & 181  \\ \hline

\end{tabular}
\label{table:sculpturesstats}
\end{table}

The remainder of this section describes the steps used to 
download,  prepare, and obtain the image pair correspondences of  the dataset. 
Additional details are given in the supplementary material.

\noindent {\bf Image extraction.}
We combine multiple sculpture datasets:~\cite{Arandjelovic11_mine,Arandjelovic12a_mine,Fouhey16,Knapitsch2017} and download  additional images from the web.

\noindent {\bf Obtaining segmentations.}
To segment
the images, RefineNET~\cite{Lin16} is trained on 2000 hand-labelled
sculptures by artists \emph{Rodin} and \emph{Henry Moore}.  It
achieves a $0.94$ IoU score and $0.97$ accuracy on the validation
dataset.  This is used for a wide variety of images and it generalises
well to new sculptures. 

\noindent {\bf Obtaining correspondences.}
\label{sec:matching}
The final step is to determine
a valid set of correspondences.  The OpenMVG
pipeline~\cite{openMVG} is used to extract an initial dense list of
correspondences between pairs of images.
The segmentation from RefineNET above  is then used
to mask out correspondences from the irrelevant background parts of the image.   
Additionally those correspondences that do not satisfy the affine fundamental matrix, which is computed using RANSAC, are removed.
Finally, those image pairs that can be mapped  by an
affine homography (i.e.\ a 2D transformation between images) are thrown out, as they will not
provide a constraint on 3D structure. 

 Despite these post-processing steps, there will still be noise in the correspondences, motivating the use of a robust  cost in our losses explained in~\secref{sec:approach}.

\section{Experiments}
\label{sec:experiments}
A challenge of our framework is to determine its prediction quality, as there is {\em no} ground truth depth information for the automatically collected \sculpdata.
To this end, LiftNet is evaluated in multiple environments and scenarios.
\emph{First}, we use a realistic synthetic dataset of sculptures SketchFab~\cite{Wiles17a} and ShapeNet~\cite{shapenet2015}  for which we can determine ground truth information and thereby correspondences between views; these are introduced below. 
LiftNet is then trained using these generated correspondences and compared to a baseline trained to explicitly regress depth on \secref{res:shapenet} and \secref{res:sketchfab}. 
\emph{Second}, we train LiftNet on real data: the \sculpdata.
This network is then compared to a number of self-supervised and supervised methods in \secref{sec:ressketchfabtrasculp}.
This evaluation is performed on two datasets: first it is performed on \scanned, a dataset of scanned objects.
Second, the evaluation is performed on SketchFab (despite the domain gap between real and synthetic images the network generalises to this new domain).
Finally, it is evaluated qualitatively on the \sculpdata \  in \secref{res:qualitative}.

\subsection{Datasets, evaluation metrics, and baselines}
\noindent \paragraph{\bf The SketchFab and ShapeNet datasets.}
SketchFab is a large dataset of synthetic 3D models of sculptures generated using photogrammetry.
There are 425 sculptures divided into 372/20/33 train/val/test sculptures.
ShapeNet consists of multiple semantic classes, each of which is divided into train/val/test using the given splits.
For evaluation, five views of each SketchFab object and $10$ views of each ShapeNet  object are rendered in Blender~\cite{Blender17} using orthographic projection and the ground truth depth extracted.
The SketchFab objects are viewed with azimuth $\in [0^{\circ}, 120^{\circ}]$, elevation $0^{\circ}$ whereas ShapeNet objects are viewed with azimuth $\in [0^{\circ}, 360^{\circ}]$ and elevation $\in [-45^{\circ}, 45^{\circ}]$.
As the depth and cameras of the renders are known, the ground truth correspondences between images can be determined by projecting the depth in the \srcview \ into the \tgtview.

\noindent \paragraph{\bf \scanned.}
Additional data is collected from the $80$ sculpture videos of~\cite{Choi2016}.
These are taken `in-the-wild' with a hand-held camera.
Of these videos, 11 objects are chosen and the sculpture region segmented.
This gives 208 images for testing.

\noindent \paragraph{\bf Evaluation metrics.}
The results are reported using multiple metrics: the $L_1$ error, root mean squared error, relative $L_1$ error,  and squared rel.~difference~\cite{Eigen14}.
To evaluate the depth prediction, it is necessary to take into account the ambiguity in the $z$ axis (the depth prediction).
This is done by allowing for a scaling/translation in depth.
Thus for all models (including those trained on ground truth depth), when reporting results, the depth prediction $d_{\text{pred}}$ for an image is first normalised by $d^{*}_{\text{pred}} = \alpha ( d_{\text{pred}} - \beta_1) + \beta_2 $ where $\beta_1$ is the median of $d_{\text{pred}}$, $\beta_2$ is the median of $d_{\text{gt}}$ and $\alpha$ allows for a scaling in depth: $\alpha = \sum_{xy} (d_{\text{pred}} * d_{\text{gt}}) / \sum_{xy} (d_{\text{pred}}^2)$. ($d_{gt}$ denotes ground truth and $\sum_{xy}$ denotes summation over pixel locations.)

\noindent \paragraph{\bf Baselines.}
In the evaluation on synthetic data, we compare against a supervised baseline, explicitly trained to regress depth.
We use the same network (e.g.\ pix2pix) as LiftNet.
The MSELoss is used but after first accounting for a scaling and translation in depth as follows.
If the depth predicted is $d_{\text{pred}}$ then the normalised depth is $d^{*}_{\text{pred}} = \alpha ( d_{\text{pred}} - \beta_1) + \beta_2 $, which is computed as described above for the evaluation metrics.
The loss is then $|d^{*}_{\text{pred}} - d_{gt}|_2$.

\subsection{Training}
\label{sec:trainingtesting}
The network is trained as follows.
Two images with correspondences are sampled  from the dataset; one is designated {\em source}, the other {\em target}.
The  \srcview \ is then input to LiftNet, which predicts the depth at all pixels.
The predicted depth of the foreground pixels $d_i$, concatenated with the $x_i$,$y_i$ position of the pixel in the image give the 3D points in the \srcview \ $\vaz X_i = [x_i, y_i, d_i, 1]^T$.
The correspondence loss --  $\mathcal{L}_{\text{corr}}$ -- is then imposed on these 3D points.

\begin{figure}
\centering
\includegraphics[width=\linewidth]{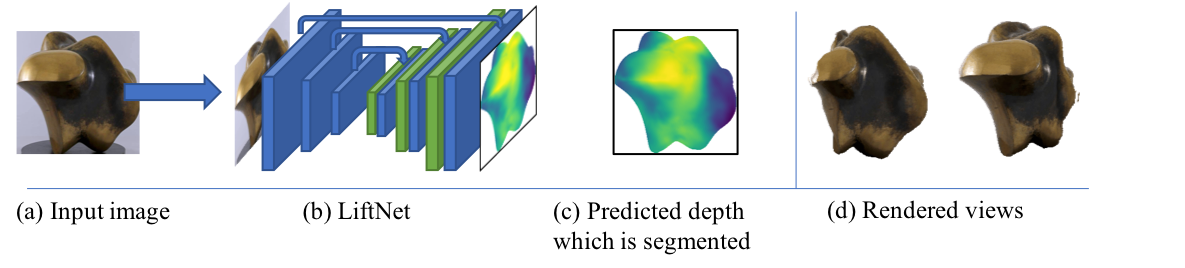}
\caption{ The test  time pipeline for LiftNet. (a) An image is selected from the test dataset and input to LiftNet (\secref{sec:architecture}). (b) LiftNet gives a depth map prediction at all points. (c-d) The rendered depth is then segmented and visualised at new views. (This is a sample result on the test set.) }
\label{fig:trainingsetup}	
\end{figure}

At {\em test} time (visualised in \figref{fig:trainingsetup}), an image is simply input to the network.
This gives the depth prediction for all pixels.
For visualisation purposes, the sculpture (the foreground pixels) are segmented from the background and only the depth values at these foreground pixels is shown.

The models  are trained on a single Titan GPU in PyTorch~\cite{Pytorch}.
They take about half a day to train.
All models trained on the \sculpdata \ are trained as follows.
The models are trained with SGD, a learning rate of $1e^{-5}$, and momentum of $0.9$.
The gradients are clamped to $\pm 5$.
These models are trained until the correspondence error on the \sculpdata's validation set stops decreasing.
When trained on  SketchFab or ShapeNet, models are trained with SGD a learning rate of $1e^{-3}$, and momentum of $0.9$.
The gradients are clamped to $\pm 5$.
They are trained until the correspondence error on the validation set stops decreasing or a maximum of $200$ epochs. 

\subsection{Quantitative results on ShapeNet}
\label{res:shapenet}
In this section, we evaluate LiftNet on ShapeNet.
In order to test the correspondence loss, $50$ correspondences per pair of images of an object are randomly chosen and fixed using the known depth and camera transformation. 
This gives the training set.

The results are reported in \tabref{shapenet} and LiftNet is compared to training the same network architecture (i.e.\ pix2pix) but directly regressing the ground truth depth up to a scaling and translation in depth as described above.

\begin{table}[]
\tiny
\centering
\caption{\small 
Comparison of `LiftNet trained on ShapeNet correspondences' to `pix2pix trained using a MSE loss on ShapeNet'.
The error measure is RMSE (x100).
}
\label{shapenet}
\begin{tabular}{l | ccccccccccccccccccc}
& rif. & boo. & bus & bed & spe. & cab. & lam.  
& cha. & tra. & pla. & tab. & dis. & mot. & car & wat. & pho. & sofa \\ \hline
pix2pix & {\bf 1.71} & {2.14} & {\bf 1.89} & {2.16} & 1.66 & 1.76 & 1.44  & 1.87 & {\bf 1.71} & {\bf 0.90} & 2.52 & 2.28 & {\bf 1.71} & 1.33 & {\bf 1.36} & {1.56} & 2.19 \\ \hline
LiftNet & {2.03} & {\bf 1.94} & {2.11}  & {\bf 1.21} & {\bf 1.29} & {\bf 1.21}  & {\bf 1.38}  & {\bf 0.94} & {2.06} & {1.05} & {\bf 1.51} & {\bf 1.66} & {1.92} & {\bf 1.12} & {1.51} & {\bf 1.52} & {\bf 1.34} \\ \hline
\end{tabular}
\end{table}
These results are perhaps surprising, as LiftNet does better on multiple classes and comparably on most.
Thus, training with a limited number of correspondences can yield comparable results to training with dense depth.

\subsection{Quantitative results on SketchFab}
\label{res:sketchfab}
In this section, LiftNet is evaluated on a synthetic dataset of sculptures, SketchFab, which has more varied shapes than ShapeNet.
LiftNet is
trained using ground truth correspondences for SketchFab for every pixel 
(i.e.\ dense points). LiftNet's performance is then compared with
the  baseline methods trained with depth.  
As demonstrated in
\tabref{results:sculpturesdataset},
our method performs similarly to the supervised method trained explicitly to regress depth.
Qualitative results are given in the supplementary material.

While here we have used all points, for ShapeNet only $50$ correspondences was sufficient.
Consequently, we additionally investigate in~\tabref{tab:robustnesscorr} the performance as a function of the number of training correspondences used per image and demonstrate that using a fraction of the available number of correspondences gives comparable results to using all. 
For example, using $100$ correspondences gives similar results -- $0.175/0.254$ L1/RMSE error versus $0.175/0.255$; we can use $1.1\%$ of the correspondences and achieve comparable results to using all. 

\begin{table}
\tiny
\centering
\caption{Evaluation of LiftNet's robustness to the number of training correspondences. Lower is better. These results demonstrate that using only 50 correspondences per training pair gives similar results to using all. Thus, sparse correspondences are sufficient for training LiftNet.}
\begin{tabular}{ c || c | c | c | c }
\# Correspondences per Image  & $L_1$ & RMSE & $\frac{d^{*} - d_{gt}}{d_{gt}}$  & $\frac{(d^{*} - d_{gt})^2}{d_{gt}}$ \\ \hline
10  & 0.183 & 0.263 & 0.0673 & 0.0253 \\ 
50  & 0.178 & 0.261 & 0.0650 & 0.0242 \\ 
100  & 0.175 & 0.254 & 0.0640 & 0.0233 \\ 
$\approx$9000  & 0.175 & 0.255 & 0.0641 & 0.0233 \\ 
 
 \end{tabular}	

\label{tab:robustnesscorr}
\end{table}

\subsection{Quantitative results using real world data}
\label{sec:ressketchfabtrasculp}
Given the initial experiments on ShapeNet and SketchFab, which demonstrate that our loss is sufficient to learn about 3D and that using sparse correspondences is powerful, we turn our attention to using real-world, noisy data.
The model is trained on the real-world images from the \sculpdata.
However, as there is no large dataset of ground truth 3D sculptures, we evaluate on two datasets.
First, we evaluate on real images using the \scanned \ dataset.
Second we evaluate the model's generalisation capabilities by evaluating on SketchFab.
To perform well, the model must generalise to a new, synthetic domain which may require a challenging domain shift.
However, in practice, the model seems robust enough to generalise to this domain.

\noindent {\bf Training.}
When training, the loss on the validation set decreases from $\approx 4.0$ to $\approx 3.4$, converging in $40K$ iterations.

\begin{table}
\centering
\caption{The performance of LiftNet evaluated on the SketchFab dataset. Across all metrics, lower is better.}
\tiny
\begin{tabular}{| P{3cm} | c | c  || c | c | c | c | }
\hline
Method & Trained with & Training Dataset &  $L_1$ & RMSE & $\frac{d^{*} - d_{gt}}{d_{gt}}$  & $\frac{(d^{*} - d_{gt})^2}{d_{gt}}$ \\ \hline \hline
{COLMAP}~\cite{schoenberger2016sfm} & Depth from SfM & Sculptures & 0.195 & 0.284 & 0.0760 & 0.0291  \\ 
{LiftNet:} $\mathcal{L}_{\text{Corr}}$ (no $R$) & Correspondences & {Sculptures}   & 0.190 & 0.277 & 0.0690 & 0.0269 \\ 
LiftNet: $\mathcal{L}_{\text{Corr}}$  & Correspondences & Sculptures & {\bf 0.186} & {\bf 0.270} & {\bf 0.0677} & {\bf 0.0256} \\ 
Zhou et.\ al.~\cite{Zhou17} & Photometric Error & Sculptures & 0.202 & 0.291 & 0.0732 & 0.0297  \\ \hline \hline
Chen {\it et al.}~\cite{Chen16} & Ground Truth Ordinal Depth & {Depth-in-the-Wild} & 0.186 & {0.269} & 0.0680 & 0.0258 \\ \hline \hline
{ LiftNet: } $\mathcal{L}_{\text{Corr}}$ & Correspondences & SketchFab & {0.175} & \bf{0.254} & 0.0641 & {0.0233} \\  
{ pix2pix} & Depth & SketchFab & \bf{0.173} & {\bf 0.254} & \bf{0.0628} & \bf{0.0226} \\ \hline
\end{tabular}
\label{results:sculpturesdataset}

\end{table}

\noindent {\bf Ablation Studies.}
The first step is to ensure that our loss does indeed enforce that LiftNet learns about depth.
To perform this check, we evaluate LiftNet on the test set of SketchFab and evaluate the effect of adding each component: the correspondence loss $\mathcal{L}_{\text{corr}}$ and the robustness term $R$.

The results are reported in \tabref{results:sculpturesdataset}.
From these results, it is clear that the correspondence loss provides a strong constraint on the predicted depth, which is improved by the robustness term.

\noindent {\bf Comparison to SfM.}
The benefit of our approach is that we do not require videos of the same object but instead can use unordered image collections and a small number of images per object.
To demonstrate this, we compare to COLMAP~\cite{schoenberger2016sfm}.
COLMAP is run on the clusters and the recovered 3D used to train a model to explicitly regress depth.
COLMAP failed for 77\% of the clusters, as there are not sufficient images/correspondences for it to converge to a global solution.
\tabref{results:sculpturesdataset} and \ref{results:scanneddataset} compares the performance of the two methods. 
The proposed pipeline and LiftNet training are superior, due to (we assume):
(1) more data for training, as no correspondences are thrown out; and (2)
that the depth from COLMAP may be incorrect due to the small number of images per cluster, which may lead to an incorrect solution.

This experiment suggests that our method is additionally useful when fine-tuning a pre-trained model (e.g.\ with ground truth depth) on a new domain with only a few images per instance (e.g.\ lesser known landmarks, sculptures, etc.) as a SfM approach would fail given the sparse amount of information.

\begin{table}
\centering
\caption{The performance of LiftNet evaluated on the \scanned \ dataset. Across all metrics, lower is better.}
\tiny
\begin{tabular}{| P{3cm} | c | c  || c | c | c | c | }
\hline
Method & Trained with & Training Dataset &  $L_1 (cm)$ & RMSE (cm) & $\frac{d^{*} - d_{gt}}{d_{gt}}$  & $\frac{(d^{*} - d_{gt})^2}{d_{gt}}$ \\ \hline \hline
{COLMAP}~\cite{schoenberger2016sfm} & Depth from SfM & Sculptures & 9.5 & 11.8 & {\bf 0.0741} & 18.1  \\ 
LiftNet: $\mathcal{L}_{\text{Corr}}$  & Correspondences & Sculptures & {\bf 9.4} & {\bf 11.6} & {\bf 0.0741} & {\bf 16.6} \\ 
Zhou et.\ al.~\cite{Zhou17} & Photometric Error & Sculptures & 9.8 & 12.1 & 0.0761 & 18.7  \\ \hline \hline
Chen {\it et al.}~\cite{Chen16} & Ground Truth Ordinal Depth & {Depth-in-the-Wild} & 9.3 & 11.7 & 0.0722 & 17.1 \\ \hline 

\end{tabular}
\label{results:scanneddataset}

\end{table}

\noindent {\bf Comparison to other self-supervised approaches.}
The second hypothesis to test is whether our method is more robust than other
self-supervised methods which rely on photometric consistency.
We compare to the work of~\cite{Zhou17} by running their model on our dataset.
However, we note that their method requires knowledge of the intrinsic camera parameters which we do not have.
As a result, we assume the intrinsic camera parameters  
have focal length 0.7*W, and the principal point is (0.5W, 0.5H) (W/H are the width/height of the image).
The results are reported in \tabref{results:sculpturesdataset}.
As can be seen their model does poorly: this is presumably due to a number of challenging characteristics of the \sculpdata.
First, as mentioned above the intrinsic camera parameters are not known and may change from image to image.
Second, there are large changes in illumination, changes in context, changes in weather, etc.
All of these characteristics make using a photometric loss not robust and lead to worse results.

\noindent {\bf Comparison to supervised approaches.}
Despite LiftNet doing better than comparable self-supervised approaches, as reported above, the next question is how does LiftNet compare to a method~\cite{Chen16} trained with depth supervision.
\cite{Chen16} is pre-trained on the NYU depth dataset~\cite{Silberman12_mine} which contains 795 densely annotated images and fine-tuned on the depth-in-the-wild \cite{Chen16} which contains 5M images with ordinal relationships. 
As demonstrated in~\tabref{results:sculpturesdataset} and \ref{results:scanneddataset}, LiftNet does comparably or better than this {\em supervised} baseline. 

\def\spacingletters{\vspace{4em}}

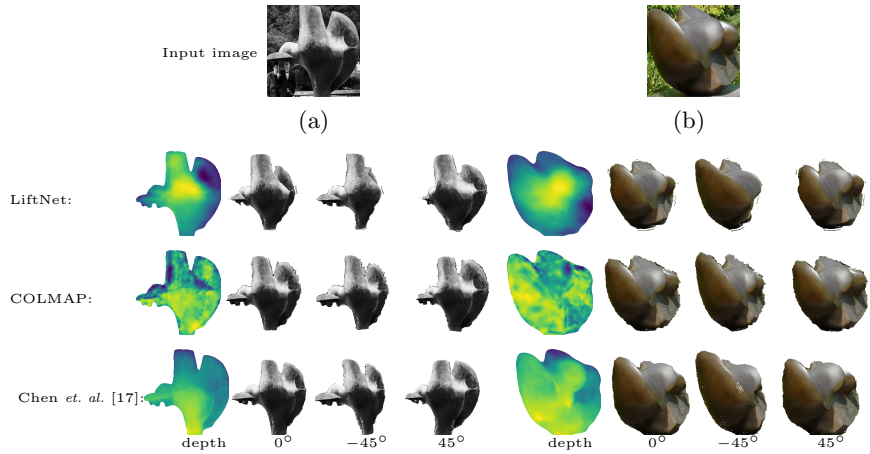
\begin{figure}
\centering
\begin{minipage}[h]{0.4\linewidth}
\centering
\subfloat[]{
\centering
\put(-40,15){\tiny Input image}

\includegraphics[width=0.25\linewidth]{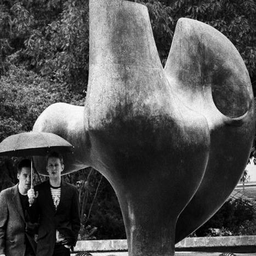}}

\vspace{0.5em}

\includegraphics[width=0.23\linewidth]{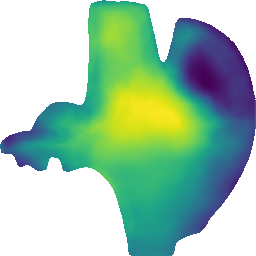}
\put(-80,12){\tiny LiftNet: }
\includegraphics[width=0.23\linewidth,trim={4.5cm .8cm 4.5cm .8cm},clip]{{{gifs15_preddepth.gif-0}.png}}
\includegraphics[width=0.23\linewidth,trim={4.5cm .8cm 4.5cm .8cm},clip]{{{gifs15_preddepth.gif-2}.png}}
\includegraphics[width=0.23\linewidth,trim={4.5cm .8cm 4.5cm .8cm},clip]{{{gifs15_preddepth.gif-10}.png}}

\vspace{0.5em}

\includegraphics[width=0.23\linewidth]{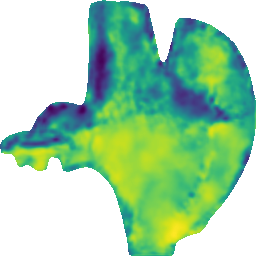}
\put(-80,12){\tiny COLMAP: }
\includegraphics[width=0.23\linewidth,trim={4.5cm .8cm 4.5cm .8cm},clip]{{{gifscolmap15_preddepth.gif-0}.png}}
\includegraphics[width=0.23\linewidth,trim={4.5cm .8cm 4.5cm .8cm},clip]{{{gifscolmap15_preddepth.gif-2}.png}}
\includegraphics[width=0.23\linewidth,trim={4.5cm .8cm 4.5cm .8cm},clip]{{{gifscolmap15_preddepth.gif-10}.png}}

\vspace{0.5em}

\includegraphics[width=0.23\linewidth]{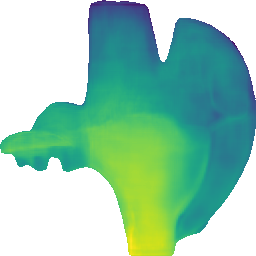}
\put(-80,12){\tiny Chen {\it et.\ al.} \cite{Chen16}: }
\includegraphics[width=0.23\linewidth,trim={4.5cm .8cm 4.5cm .8cm},clip]{{{gifschenetal15_preddepth.gif-0}.png}}
\put(-15,-5){\tiny $0^{\circ}$}
\put(-50,-5){\tiny depth }
\includegraphics[width=0.23\linewidth,trim={4.5cm .8cm 4.5cm .8cm},clip]{{{gifschenetal15_preddepth.gif-2}.png}}
\put(-20,-5){\tiny $-45^{\circ}$}
\includegraphics[width=0.23\linewidth,trim={4.5cm .8cm 4.5cm .8cm},clip]{{{gifschenetal15_preddepth.gif-10}.png}}
\put(-17,-5){\tiny $45^{\circ}$}
\vspace{0.5em}
\end{minipage}
\begin{minipage}[h]{0.4\linewidth}
\centering
\subfloat[]{
\centering
\includegraphics[width=0.25\linewidth]{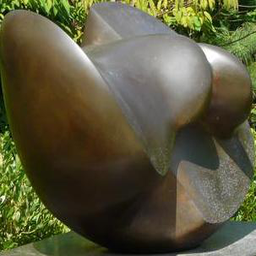}}

\vspace{0.5em}

\includegraphics[width=0.23\linewidth]{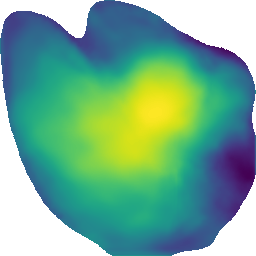}
\includegraphics[width=0.23\linewidth,trim={4.5cm .8cm 4.5cm .8cm},clip]{{{gifs860_preddepth.gif-0}.png}}
\includegraphics[width=0.23\linewidth,trim={4.5cm .8cm 4.5cm .8cm},clip]{{{gifs860_preddepth.gif-3}.png}}
\includegraphics[width=0.23\linewidth,trim={4.5cm .8cm 4.5cm .8cm},clip]{{{gifs860_preddepth.gif-9}.png}}

\vspace{0.5em}

\includegraphics[width=0.23\linewidth]{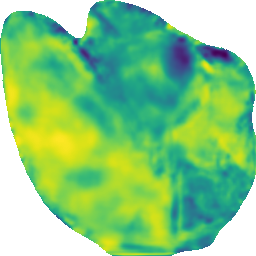}
\includegraphics[width=0.23\linewidth,trim={4.5cm .8cm 4.5cm .8cm},clip]{{{gifscolmap860_preddepth.gif-0}.png}}
\includegraphics[width=0.23\linewidth,trim={4.5cm .8cm 4.5cm .8cm},clip]{{{gifscolmap860_preddepth.gif-3}.png}}
\includegraphics[width=0.23\linewidth,trim={4.5cm .8cm 4.5cm .8cm},clip]{{{gifscolmap860_preddepth.gif-9}.png}}

\vspace{0.5em}

\includegraphics[width=0.23\linewidth]{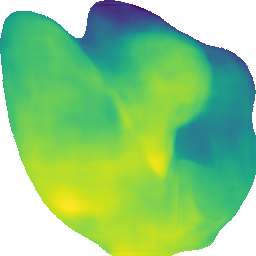} 
\includegraphics[width=0.23\linewidth,trim={4.5cm .8cm 4.5cm .8cm},clip]{{{gifschenetal860_preddepth.gif-0}.png}}
\put(-17,-5){\tiny $0^{\circ}$}
\put(-55,-5){\tiny depth }
\includegraphics[width=0.23\linewidth,trim={4.5cm .8cm 4.5cm .8cm},clip]{{{gifschenetal860_preddepth.gif-3}.png}}
\put(-23,-5){\tiny $-45^{\circ}$}
\includegraphics[width=0.23\linewidth,trim={4.5cm .8cm 4.5cm .8cm},clip]{{{gifschenetal860_preddepth.gif-9}.png}}
\put(-17,-5){\tiny $45^{\circ}$}
\vspace{0.5em}
\end{minipage}
\caption{
Reconstruction results for LiftNet (top), COLMAP (middle) and Chen {\it et al.}~\cite{Chen16}
(bottom), visualised
using Open3D~\cite{Zhou2018}. The input image is shown at the top, then the
predicted depth (blue is further away, yellow nearer), and rendered
3D at multiple viewpoints.  Zoom in
for details.
From these images, the following are demonstrated.
First,  Chen {\it et al.} learns a prior over the image that the bottom of the image is nearer and the top further away.
This is demonstrated in (a) and further examples in the supp.~material. 
Second, COLMAP's depth predictions are noisy.
Finally, LiftNet produces convincing depth maps which can be rendered at new views. 
}
\label{fig:qualitative_sculpture_results}	
\end{figure}

\subsection{Qualitative Results on the \sculpdata}
\label{res:qualitative}
We demonstrate in \figref{fig:qualitative_sculpture_results} the
predictions of LiftNet on the testing portion of the \sculpdata \ and compare them visually to two other methods: COLMAP and the supervised method~\cite{Chen16}.
We note
that COLMAP performs poorly, presumably as there are very few training points.
\cite{Chen16} produces reasonable results, as it is trained on a large dataset of outdoors images with supervision on relative depth in addition to NYU, but it has certain priors over the image (e.g. that points in the bottom of the image are always nearer than those in the top -- as for most images the foreground is at the bottom of the image and sky at the top).
Please see the supplementary material for more results.

\section{Discussion}
\label{sec:conclusions}
In this paper, we have introduced a framework for learning 3D shape using easily attainable sparse correspondences {\em without} depth supervision.  
Our insight is that we can make use of sparse correspondences, which can be obtained in much less constrained environments than approaches requiring photometric consistency.
Given enough sparse correspondences across many instances, the network learns a dense depth prediction.
The approach  has been
demonstrated on a challenging sculpture dataset of real images and a synthetic sculpture dataset with known ground truth information.

It is interesting to consider why this training scenario based on real
images, and sculptures in particular, produces a network that performs
well on real images and also generalizes to synthetic image. It is
probably in part because the training data has natural augmentation --
instances of a sculpture with the same shape may be made from
different materials (bronze, marble) or have different texturing
and appearance due to different weathering or illumination conditions. The
network must learn to produce the same shape, irrespective of these
multifarious conditions. This is a challenging learning problem but,
if successful,  then the network has correctly learnt to disentangle the
material/appearance from the shape, and to pick out cues to shape from
 appearance. Thus it can  generalize to objects with different
materials, e.g.\ synthetic ones.

\paragraph{Acknowledgements.}
The authors would like to thank Fatma Guney for helpful feedback and suggestions.
This work was funded by an EPSRC studentship and EPSRC Programme
Grant Seebibyte EP/M013774/1.

\clearpage

\bibliographystyle{splncs}
\bibliography{shortstrings,vgg_other,vgg_local,my_citations}
\end{document}